\documentclass[letterpaper]{article} 
\usepackage{aaai2026}  
\usepackage{times}  
\usepackage{helvet}  
\usepackage{courier}  
\usepackage[hyphens]{url}  
\usepackage{graphicx} 
\usepackage{graphicx} 
\usepackage{natbib}  
\usepackage{caption} 
\usepackage{algorithm}
\usepackage{algorithmic}
\usepackage{amsmath}   
\usepackage{multirow} 

\usepackage{newfloat}
\usepackage{listings}
\DeclareCaptionStyle{ruled}{labelfont=normalfont,labelsep=colon,strut=off} 
\floatstyle{ruled}
\newfloat{listing}{tb}{lst}{}
\floatname{listing}{Listing}
\title{ALEX:A Light Editing-knowledge Extractor}
\author {
    Minghu Wang\textsuperscript{\rm 1,2,3},
    Shuliang Zhao\textsuperscript{\rm 1,2,3}\thanks{Corresponding author},
    Yuanyuan Zhao\textsuperscript{\rm 2,3,4,5},
    Hongxia Xu\textsuperscript{\rm 1,2,3}
}

\affiliations {
    \textsuperscript{\rm 1}College of Computer and Cyber Security, Hebei Normal University, Hebei 050024, China\\
    \textsuperscript{\rm 2}Hebei Provincial Engineering Research Center for Supply Chain Big Data Analytics \& Data Security, Hebei 050024, China\\
    \textsuperscript{\rm 3}Hebei Provincial Key Laboratory of Network and Information Security, Hebei 050024, China\\
    \textsuperscript{\rm 4}School of Mathematical Sciences, Hebei Normal University, Hebei 050024, China\\
    \textsuperscript{\rm 5}Dept of Information Engineering, Shijiazhuang College of Applied Technology, Hebei 050043,  China\\
}

\usepackage{bibentry}

\begin{document}

\maketitle

\begin{abstract}

The static nature of knowledge within Large Language Models (LLMs) makes it difficult for them to adapt to evolving information, rendering knowledge editing a critical task. However, existing methods struggle with challenges of scalability and retrieval efficiency, particularly when handling complex, multi-hop questions that require multi-step reasoning. To address these challenges, this paper introduces ALEX (A Light Editing-knowledge Extractor), a lightweight knowledge editing framework. The core innovation of ALEX is its hierarchical memory architecture, which organizes knowledge updates (edits) into semantic clusters. This design fundamentally reduces retrieval complexity from a linear 
$O(N)$ to a highly scalable $O(K+N/C)$. Furthermore, the framework integrates an Inferential Query Synthesis (IQS) module to bridge the semantic gap between queries and facts , and a Dynamic Evidence Adjudication (DEA) engine that executes an efficient two-stage retrieval process. Experiments on the MQUAKE benchmark demonstrate that ALEX significantly improves both the accuracy of multi-hop answers (MultiHop-ACC) and the reliability of reasoning paths (HopWise-ACC). It also reduces the required search space by over 80\% , presenting a promising path toward building scalable, efficient, and accurate knowledge editing systems.

\end{abstract}

\section{Introduction}

Large Language Models (LLMs) have demonstrated remarkable capabilities in natural language understanding and generation, serving as extensive repositories of world knowledge. However, a fundamental limitation of these models is the static nature of their internal knowledge, which is acquired during a resource-intensive pre-training phase. As the world evolves and new information emerges, LLMs can quickly become outdated, leading to the generation of factually incorrect or ``hallucinated'' responses. The critical task of updating, correcting, or extending the knowledge within these models without resorting to complete retraining is known as knowledge editing~\cite{gu-etal-2024-PokeMQA,meng2022locating,meng2023massediting,zhong-etal-2023-MQuAKE}.

\begin{figure}[t]

\centering

\includegraphics[width=0.9\columnwidth]{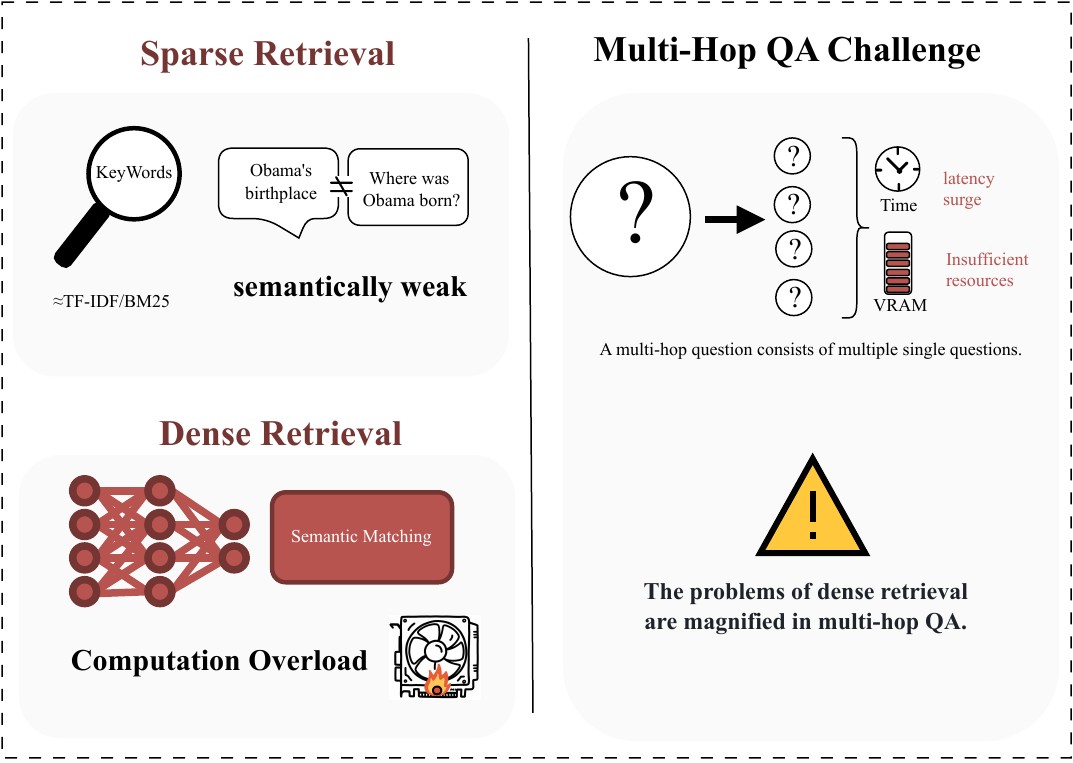} 

\caption{Sparse vs. Dense Retrieval: Speed vs. Accuracy.}

\label{intro}

\end{figure}

Current approaches to knowledge editing predominantly fall into two categories: parametric and memory-based methods. Parametric methods, such as ROME~\cite{meng2023massediting} and MEMIT~\cite{meng2022locating}, directly modify the model's internal weights to instill new factual associations. While precise, these techniques often suffer from unintended consequences; editing one fact can interfere with related information, causing cascading errors that disrupt the model's reasoning capabilities, especially in complex multi-hop scenarios. In contrast, memory-based approaches like MeLLo~\cite{zhong-etal-2023-MQuAKE} and PokeMQA~\cite{gu-etal-2024-PokeMQA} store factual updates in an external memory, decoupling the editing process from the model's core parameters. This offers greater flexibility but introduces significant challenges in scalability and retrieval efficiency~\cite{zhang-etal-2024-end,zhuang-etal-2024-efficientrag}.

These challenges are magnified in the context of multi-hop question answering, which requires models to decompose complex queries and synthesize information across multiple reasoning steps. The effectiveness of memory-augmented systems in this domain hinges on their ability to accurately and efficiently retrieve the correct edit at each step. However, a persistent ``semantic gap'' exists between the phrasing of user queries and the declarative format of stored edits, making retrieval fragile. This reflects a well-known trade-off: efficient sparse retrieval methods~\cite{jiang-etal-2023-active,trivedi-etal-2023-interleaving} often fail to capture semantic nuances~\cite{nguyen-etal-2024-dyvo,lee-etal-2023-complementarity}, while more accurate dense retrievers incur substantial computational overhead~\cite{reichman-heck-2024-dense,gao-etal-2023-precise}. While strategies like Chain-of-Thought (CoT) have improved multi-step reasoning over static knowledge, they lack native mechanisms to incorporate external updates, often leading to error propagation when underlying facts change. This highlights a critical need for a knowledge editing framework that is simultaneously scalable, efficient, and robust to the semantic diversity of natural language queries.

To address these limitations, we propose ALEX (A Light Editing-knowledge Extractor), a novel framework that combines hierarchical memory compression with dynamic retrieval DEA to achieve accurate and efficient knowledge editing at scale. Crucially, as its name suggests, ALEX is designed not as a standalone editor but as a lightweight, modular extractor that can be seamlessly integrated into existing memory-based methods to replace their inefficient retrieval components. ALEX introduces a hierarchical architecture that transforms the edit memory from a flat, linear structure into a network of semantically organized clusters, using the K-means++ algorithm~\cite{hassan2025hybrid} for its strong empirical performance and minimal overhead. This design fundamentally changes the retrieval complexity from a linear $O(N)$ to a highly scalable $O(K + N/C)$. Our framework is composed of three synergistic modules:

1. The \textbf{Semantic Manifold Partitioning (SMP) Engine} proactively organizes edits into semantically cohesive clusters, trained with a dual-objective that optimizes intra-cluster cohesion and inter-cluster contrastiveness~\cite{ding2025clustering}.

2. The \textbf{Inferential Query Synthesis (IQS) Module} bridges the semantic gap by reformulating each stored fact into a diverse set of hypothetical questions, anticipating how a user might inquire about that information.

3. The \textbf{Dynamic Evidence Adjudication (DEA) Engine} implements an efficient two-stage retrieval process that first identifies relevant clusters and then pinpoints the most salient edit using both direct and inferential matching signals.

In summary, the contributions of our paper are as follows:

\begin{itemize}

\item A dual-objective training strategy unifies contrastive clustering with QA ranking, enabling rapid convergence (5 epochs) while preserving semantic fidelity across clusters.

\item The \textbf{Inferential Query Synthesis (IQS)} and \textbf{Dynamic Evidence Adjudication (DEA)} modules collaborate to optimize retrieval. This synergy enhances matching precision through pseudo-query scoring and reduces the edit search space by over 80\% through hierarchical filtering.

\item ALEX achieves state-of-the-art performance on MQuAKE, demonstrating effectiveness in both accuracy and scalability.

\end{itemize}

\begin{figure*}[t]
\centering
\large
\includegraphics[width=0.8\textwidth]{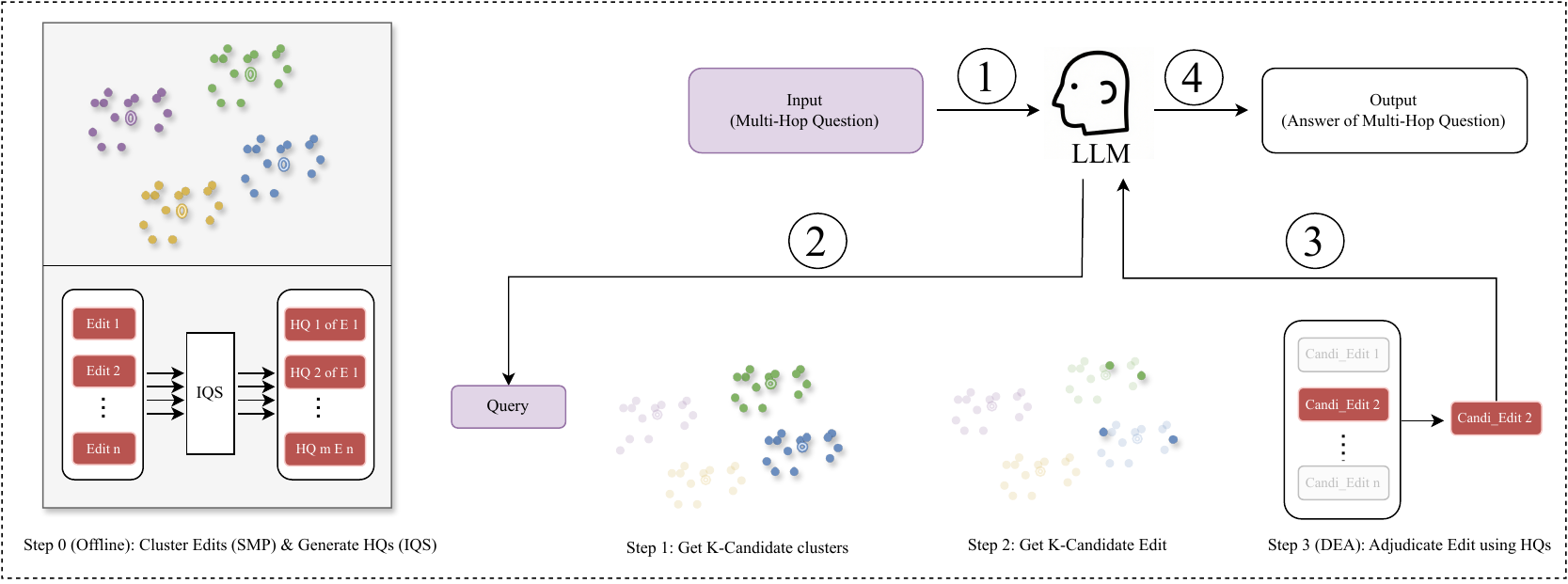} 
  \caption{Workflow of our knowledge editing system.}
\label{frame}
\end{figure*}

\section{Related Work}

\paragraph{Knowledge Editing Techniques}

Contemporary approaches to knowledge editing in large language models bifurcate into parametric and memory-based methodologies. An ``edit'' itself is the specific factual update, often expressed as a declarative statement (e.g., ``The Eiffel Tower is located in Paris'') intended to correct or add to the model's knowledge.

Parametric methods, such as ROME and MEMIT, focus on modifying internal model weights to update factual associations. While ROME locates specific layers for targeted edits, MEMIT extends this capability to handle mass edits through multi-layer adjustments. However, these methods inherently suffer from cascading interference—editing one fact often disrupts related knowledge structures, leading to inconsistent outputs in multi-hop reasoning scenarios. Moreover, their reliance on static parameter updates renders them ill-suited for dynamic environments requiring frequent knowledge updates.

In contrast, memory-based approaches like MeLLo and PokeMQA decouple knowledge storage from model parameters by maintaining edits in memory. MeLLo iteratively prompts language models to align generated answers with stored edits, while PokeMQA introduces programmable scope detectors to disentangle conflict resolution from question decomposition. Despite their flexibility, these methods face challenges in scalability: memory consumption for dense retrievers (e.g., Contriever) grows linearly with the number of edits, and full-index similarity computations incur significant computational costs, particularly for multi-hop queries. Recent studies further highlight an accuracy-efficiency trade-off in lightweight retrievers, where aggressive compression strategies often degrade semantic matching fidelity. Furthermore, the retrieval effectiveness of these systems often hinges on the direct semantic overlap between a query and the stored declarative edits, creating a vulnerability when faced with the diverse phrasings and inferential nature of multi-hop questions.

\paragraph{Efficient Retrieval Paradigms}

Dense retrieval systems, exemplified by DPR\cite{karpukhin-etal-2020-dense} and Contriever\cite{izacard2021unsupervised}, leverage transformer-based encoders to map queries and documents into continuous vector spaces. While these models excel at semantic matching, their computational overhead scales significantly with dataset size due to exhaustive similarity calculations. The storage demands of high-dimensional embeddings further exacerbate deployment challenges on resource-constrained platforms. Clustering-based optimizations, such as hierarchical indexing and product quantization, mitigate these issues by approximating nearest neighbor searches. However, such methods struggle with concept drift—static clusters fail to adapt to incremental edits, while retraining indices introduces operational latency.

Sparse retrieval systems, including BM25\cite{10.1561/1500000019} and SPLADE\cite{10.1145/3404835.3463098}, prioritize efficiency through inverted indices and term-frequency heuristics. Their interpretable keyword-matching mechanisms enable rapid document retrieval, particularly for lexically overlapping queries. Yet, their inability to handle semantic variations (e.g., paraphrases or polysemy) significantly limits performance in complex multi-hop scenarios. Hybrid architectures\cite{ren-etal-2023-retrieve,xia-etal-2024-hybrid} attempt to reconcile these trade-offs but often introduce architectural complexity, as seen in late-fusion models that separately process lexical and semantic signals.

\paragraph{Multi-Hop Reasoning Architectures}

Chain-of-Thought (CoT) prompting has become a mainstream strategy for multi-step reasoning, enabling language models to decompose complex questions into intermediate subqueries. While effective for reasoning over static knowledge, CoT lacks mechanisms to incorporate external updates, often resulting in error propagation when underlying facts change. To address this, recent memory-augmented systems introduce retrieval at each reasoning step. However, as the number of edits grows and queries become more semantically complex, ensuring both retrieval accuracy and efficiency becomes increasingly challenging. The absence of adaptive reuse mechanisms further limits responsiveness, especially in real-time settings. Collectively, these limitations highlight a critical need for a knowledge editing framework that can scale efficiently, bridge the semantic gap between user queries and stored facts, and adapt dynamically to evolving knowledge within complex reasoning workflows.

\section{ALEX}
\subsection{Framework Overview}

Existing memory-based methods for multi-hop question answering, such as MeLLo and PokeMQA, struggle with scalability due to linear memory growth and high computational costs during retrieval. To overcome these challenges, we introduce \textbf{ALEX}, a memory-efficient framework built on a novel hierarchical architecture. At its core, ALEX partitions the entire knowledge edit memory into distinct semantic clusters. This design transforms the memory complexity from a linear $O(N)$ to a much more scalable $O(K+N/C)$, where $N$ is the total number of edits, $K$ is the number of clusters, and $C$ is the average cluster size.

The framework's workflow is implemented through three synergistic modules, which are detailed in the subsequent sections:
1.  The \textbf{Semantic Manifold Partitioning (SMP) Engine} proactively organizes the edits into a structured, hierarchical memory.
2.  The \textbf{Inferential Query Synthesis (IQS) Module} bridges the lexical gap between user queries and stored facts by reformulating each edit into a set of diverse, hypothetical questions.
3.  The \textbf{Dynamic Evidence Adjudication (DEA) Engine} executes an efficient, two-stage retrieval process.

When a multi-hop question is presented, it is first decomposed into simpler sub-queries. For each sub-query, the DEA engine rapidly identifies relevant clusters and then pinpoints the most salient edit within them, leveraging the synthesized questions from the IQS module to ensure robust matching. As illustrated in Figure~\ref{frame}, this integrated architecture ensures both high retrieval efficiency and accurate multi-step reasoning.

\subsection{Semantic Manifold Partitioning (SMP) Engine}
To address the scalability and semantic limitations of standard dense retrieval in multi-hop settings, we propose a \textbf{Semantic Manifold Partitioning (SMP) Engine} that hierarchically organizes the edit space. Unlike standard retrieval methods that suffer from linear computational growth and lack structural modularity, our SMP engine partitions the edit memory into semantically cohesive and structurally diverse groups. This not only reduces search redundancy but also facilitates efficient information updating and localized retrieval.

\textbf{Dual-Objective Optimization.}
We design a training framework that jointly optimizes two objectives: promoting intra-cluster coherence and enforcing inter-cluster separability. Given a training batch, the model minimizes the following composite loss:
\begin{equation}
\label{eq:total_loss}
\mathcal{L}_{\text{train}} = \lambda \mathcal{L}_{\text{cohesion}} + (1-\lambda) \mathcal{L}_{\text{contrast}},
\end{equation}
where the cohesion loss encourages the alignment of each edit with its assigned cluster centroid:
\begin{equation}
\label{eq:intra}
\mathcal{L}_{\text{cohesion}} = -\frac{1}{K}\sum_{c=1}^{K} \frac{1}{|\mathcal{G}_c|} \sum_{e \in \mathcal{G}_c} \cos\left(\phi(e), \mu_c\right),
\end{equation}
and the contrastive loss promotes inter-cluster distinction. For each edit $e_i$ in a batch $B$, we use a corresponding synthetically generated query $q_i$ as an anchor. The loss penalizes similarity between this anchor and other edits (negatives) in the batch from different clusters:
\begin{equation}
\label{eq:contrast}
\mathcal{L}_{\text{contrast}} = -\frac{1}{B} \sum_{i=1}^{B} \log \frac{\exp\left(\cos(\phi(q_i), \phi(e_i))/\tau\right)}{\sum_{e_j \in \mathcal{N}_i} \exp\left(\cos(\phi(q_i), \phi(e_j))/\tau\right)}.
\end{equation}
Here, $\lambda$ is a hyperparameter balancing the two objectives, set to $0.4$ based on performance on a held-out validation set. $\phi(\cdot)$ denotes the embedding function, $\cos(\cdot, \cdot)$ is the cosine similarity, and $\mathcal{N}_i$ represents the set of in-batch negatives for $q_i$. The clustering model is initialized using a K-means++ scheme where initial centroids are sampled with probabilities weighted by their similarity to a diverse set of pre-selected anchor edits, improving initial cluster quality. Further training and optimization details are provided in Appendix~E.

\textbf{Multi-Modal Representation.}
To support robust partitioning across semantically diverse edits, each edit $e_i$ is encoded with a hybrid feature vector that incorporates both neural and lexical signals:
\begin{equation}
\label{eq:feature}
\mathbf{f}_i = \mathrm{concat}\left(\phi(e_i), \frac{\ell(e_i)}{L_{\max}}, \frac{w(e_i)}{W_{\max}}\right),
\end{equation}
where $\phi(e_i) \in {R}^d$ is the MPNet embedding, and $\ell(\cdot), w(\cdot)$ are the character length and word count of the edit, which are then normalized by their maximum values ${L_{\max}}$ and ${W_{\max}}$, respectively.

\textbf{Dynamic Manifold Adaptation.}
Post-training, the SMP engine supports dynamic updates. This mechanism monitors cluster cohesion via metrics such as the silhouette score and automatically triggers partial reclustering when the average silhouette score of any cluster falls below a threshold $\theta_s=0.5$, or the global average score drops by more than 20\% relative to its post-training peak. This ensures the manifold structure adapts to evolving data distributions. The detailed procedure is elaborated in Appendix~B.

\subsection{Inferential Query Synthesis (IQS) Module}
In memory-based knowledge editing, factual updates are typically stored as declarative statements. However, user queries—especially in multi-hop question answering—often differ from these statements in surface form. This discrepancy makes it difficult to retrieve relevant edits using direct query-edit similarity alone.
To bridge this gap, the \textbf{Inferential Query Synthesis (IQS)} module reformulates each edited fact $e_j$ into a set of ${N_h}$ hypothetical questions $\mathcal{H}(e_j) = \{h_1^{(j)}, h_2^{(j)}, \dots, h_{N_h}^{(j)}\}$ by prompting a large language model. These questions aim to capture diverse and plausible ways a user might inquire about the underlying fact.
To ensure quality, the generated question set $\mathcal{H}(e_j)$ is scored using a composite function that balances semantic alignment with the edit against internal diversity. The score is calculated as a trade-off between a \textbf{relevance score} and a \textbf{redundancy penalty}.
First, we define the average relevance of the hypothetical questions to the original edit $e_j$:
\begin{equation}
\label{eq:relevance}
\mathcal{R}(e_j, \mathcal{H}(e_j)) = \frac{1}{N_h} \sum_{i=1}^{N_h} \cos(\phi(h_i^{(j)}), \phi(e_j)).
\end{equation}
Next, we define the internal redundancy of the question set, measured as the average pairwise similarity between all questions:
\begin{equation}
\label{eq:redundancy}
\mathcal{D}(\mathcal{H}(e_j)) = \frac{2}{{N_h}({N_h}-1)} \sum_{1 \leq i < k \leq {N_h}} \cos(\phi(h_i^{(j)}), \phi(h_k^{(j)})).
\end{equation}
The final quality score $S_j$ for the question set is then given by:
\begin{equation}
\label{eq:IQS_score_combined}
S_j = \mathcal{R}(e_j, \mathcal{H}(e_j)) - \gamma \mathcal{D}(\mathcal{H}(e_j)),
\end{equation}
where $\phi(\cdot)$ is the sentence embedding function shared across the system. The hyperparameter $\gamma$, which controls the trade-off, was set to $0.3$ empirically on a validation set to penalize excessive redundancy without sacrificing relevance. Details on prompt templates, caching, and filtering heuristics are provided in Appendix~C.

\subsection{Dynamic Evidence Adjudication (DEA) Engine}
To robustly identify the most relevant factual edit for a given query while ensuring efficiency, we propose the \textbf{Dynamic Evidence Adjudication (DEA) Engine}. This module addresses two central challenges: (i) the computational burden of exhaustive search, and (ii) the retrieval fragility against paraphrased or compositional queries. The DEA engine alleviates these issues via a two-stage pipeline that combines statistical filtering with semantically enriched scoring.

\paragraph{Stage I: Cluster-Level Statistical Filtering.}
Given a query embedding $\phi(q)$ and a set of K cluster centroids $\{\mu_i\}_{i=1}^K$, we first compute the normalized relevance of each cluster using a z-score transformation over cosine similarity. Let:
\begin{equation}
\label{eq:zscore}
z_i = \frac{\cos(\phi(q), \mu_i) - \bar{s}}{\sigma_s},
\end{equation}
where $\bar{s}$ and $\sigma_s$ are the mean and standard deviation of the cosine similarities between the query $\phi(q)$ and all cluster centroids. Clusters with $z_i \ge \zeta$ (we use $\zeta=1.0$) are retained as candidate clusters. To ensure computational tractability, we cap the number of selected clusters at $M=3$. This value was determined on a validation set to offer a robust balance between recall and efficiency. This two-stage approach—a dynamic z-score filter followed by a fixed cap—provides a flexible yet bounded selection mechanism. The z-score acts as a coarse-grained heuristic, while the cap $M$ guarantees a predictable computational load, preventing the selection of an overly large set of clusters.

\paragraph{Stage II: Edit-Level Semantic Adjudication.}
Within the filtered clusters, each candidate edit $e_j$ is scored using a composite function that integrates two forms of evidence: direct alignment (literal evidence) and reasoning consistency (inferential evidence). The literal evidence is measured by the direct cosine similarity between the query and the edit, while the inferential evidence is evaluated by matching the query against a set of hypothetical questions $\mathcal{H}(e_j)$. The final adjudication score is defined as:

\begin{equation}
\label{eq:adj_score}
\begin{aligned}
\Psi(e_j) ={}& \alpha \cdot \cos(\phi(q), \phi(e_j)) \\
             &+ \beta \cdot \mathrm{Agg}_{h \in \mathcal{H}(e_j)}
               \cos(\phi(q), \phi(h)).
\end{aligned}
\end{equation}
The trade-off parameters $\alpha$ and $\beta$ are set to $0.5$ to give equal weight to literal and inferential evidence. We employ max-pooling for the $\mathrm{Agg}$ operator, as it proved more effective in our preliminary experiments at isolating the signal from the single most relevant hypothetical question.

\paragraph{Selection.}
The edit with the highest adjudication score is deterministically selected:
\begin{equation}
\label{eq:selectmax}
e^* = \arg\max_{e_j} \Psi(e_j).
\end{equation}
This two-stage process effectively prunes the search space. The z-score filtering discards statistically irrelevant portions of the memory, while the semantic adjudication stage uses the synthesized queries from the IQS module to ensure that even complex or paraphrased user queries can be robustly matched to the correct underlying fact. This synergy between statistical pruning and deep semantic evaluation allows the engine to achieve high efficiency without compromising retrieval fidelity.

\section{Experiments}
\subsection{Experimental Setup and Baselines}
We evaluate on five MQuAKE-related datasets \cite{zhong-etal-2023-MQuAKE,wang2024deepedit} (Appendix~D) with three LLM backbones \cite{touvron2023llama,dubey2024llama,dai-etal-2024-deepseekmoe}. Hardware and training details are in Appendix~E. We compare our retriever, ALEX, against several non-parametric baselines. To ensure a fair comparison, the hyperparameters for all baseline models were configured by following the recommendations in their respective original papers and validated on our held-out set.
\textbf{MeLLo} \cite{zhong-etal-2023-MQuAKE} is a memory-based method that decomposes multi-hop questions for external memory retrieval; its performance relies on the quality of decomposition.
\textbf{DeepEdit} \cite{wang2024deepedit} is a non-parametric, decoding-based method that employs a depth-first search strategy to build reasoning chains. At each step, it retrieves candidate facts to guide the generation process.
\textbf{PokeMQA} \cite{gu-etal-2024-PokeMQA} uses a decoupled approach with a scope detector and knowledge prompts, whose accuracy is critical for performance.

\begin{table*}[t]
\centering
\begin{tabular}{ll rr rr rr rr rr}
\hline
& &
\multicolumn{2}{c}{\textbf{M-CF-3K}} &
\multicolumn{2}{c}{\textbf{M-CF-3K-v2}} &
\multicolumn{2}{c}{\textbf{M-CF-T}} &
\multicolumn{2}{c}{\textbf{M-CF-2002}} &
\multicolumn{2}{c}{\textbf{M-CF-Hard}}\\
\textbf{LLM} & \textbf{Method} &
\textbf{MA} & \textbf{HA} &
\textbf{MA} & \textbf{HA} &
\textbf{MA} & \textbf{HA} &
\textbf{MA} & \textbf{HA} &
\textbf{MA} & \textbf{HA}\\
\hline
\multirow{6}{*}{DeepSeek}
& DeepEdit                & 37.88 & 26.02 & 39.81 & 31.74 & 80.00 & 62.95 & 48.53 & 35.57 & 49.31 & 41.13\\
& DeepEdit (ALEX)         & \textbf{41.95} & \textbf{30.12} & \textbf{44.88} & \textbf{34.96} & \textbf{84.75} & \textbf{70.23} & \textbf{54.87} & \textbf{41.98} & \textbf{54.92} & \textbf{47.96}\\
& MeLLo                   & 31.70 & 21.55 & 36.50 & 25.70 & 80.60 & 61.92 & 42.70 & 30.65 & 2.30  & 2.85\\
& MeLLo (ALEX)            & \textbf{36.40} & \textbf{28.85} & \textbf{42.00} & \textbf{33.40} & \textbf{92.70} & \textbf{81.64} & \textbf{49.10} & \textbf{40.93} & \textbf{2.60} & \textbf{3.95}\\
& PokeMQA                 & 41.90 & 29.83 & 45.00 & 32.90 & 78.80 & 56.47 & 55.50 & 38.21 & 39.00 & 25.78\\
& PokeMQA (ALEX)          & \textbf{51.33} & \textbf{43.60} & \textbf{53.50} & \textbf{47.43} & \textbf{87.33} & \textbf{76.49} & \textbf{65.58} & \textbf{58.42} & \textbf{79.20} & \textbf{74.35}\\
\hline
\multirow{4}{*}{Llama3.1 8B}
& MeLLo               &  8.36 &  3.78 &  9.16 &  4.03 & 46.78 & 61.92 & 10.08 &  4.59 &  0.69 &  1.04\\
& MeLLo (ALEX)        & \textbf{ 9.61} & \textbf{ 4.81} & \textbf{10.53} & \textbf{ 5.24} & \textbf{53.80} & \textbf{81.64} & \textbf{11.59} & \textbf{ 5.03} & \textbf{ 0.79} & \textbf{ 1.61}\\
& PokeMQA             & 34.96 & 21.39 & \textbf{41.00} & 26.16 & 82.76 & 71.47 & 44.00 & 28.34 & 24.94 & 17.21\\
& PokeMQA (ALEX)      & \textbf{40.43} & \textbf{34.23} & 39.32 & \textbf{26.76} & \textbf{85.70} & \textbf{76.01} & \textbf{46.30} & \textbf{40.60} & \textbf{37.29} & \textbf{30.76}\\
\hline
\multirow{6}{*}{Llama 2 7B}
& DeepEdit            & 11.30 &  6.45 & 13.72 &  7.61 & 38.13 & 30.71 & 12.88 & 10.06 &  7.03 &  6.17\\
& DeepEdit (ALEX)     & \textbf{14.05} & \textbf{ 8.03} & \textbf{16.04} & \textbf{ 8.52} & \textbf{44.96} & \textbf{37.98} & \textbf{14.97} & \textbf{11.98} & \textbf{ 8.48} & \textbf{ 7.47}\\
& MeLLo               & 16.10 &  8.58 & 14.50 &  6.85 & 59.26 & 32.95 & 15.68 &  8.44 &  3.26 &  0.55\\
& MeLLo (ALEX)        & \textbf{18.50} & \textbf{11.25} & \textbf{16.70} & \textbf{ 8.69} & \textbf{68.20} & \textbf{42.63} & \textbf{16.18} &  4.80 & \textbf{ 3.75} & \textbf{ 0.91}\\
& PokeMQA             & \textbf{34.20} & 22.91 & \textbf{37.20} & \textbf{24.81} & 67.55 & 53.57 & 43.80 & 30.66 & 35.89 & 23.19\\
& PokeMQA (ALEX)      & 33.43 & \textbf{36.80} & 36.80 & 24.90 & \textbf{78.53} & \textbf{64.34} & \textbf{45.70} & \textbf{31.46} & \textbf{40.56} & \textbf{28.90}\\
\hline
\end{tabular}
\caption {Performance comparison of different methods on multi-hop question answering with knowledge editing across various datasets using different language models. Here, the prefix “M-” denotes datasets that are part of the MQuAKE-related benchmark. Evaluation metrics include Multi-hop Accuracy (MA) and HopWise Answer Accuracy (HA).}
\label{mainResults}
\end{table*}

\subsection{Evaluation Metrics}
We evaluate model performance using \textbf{MultiHop-ACC} (MA) and \textbf{HopWise-ACC} (HA) \cite{zhong-etal-2023-MQuAKE,saxena-etal-2020-improving,chen-etal-2020-hybridqa,fei-etal-2022-cqg}. To assess our retriever's effectiveness, we measure \textbf{Cluster ACC} (identifying the correct edit's cluster) and \textbf{Retrieval ACC} (retrieving the correct edit from the cluster) \cite{gaido-etal-2024-speech,li-etal-2024-optimizing,chen-etal-2023-dsee,cui-sachan-2023-adaptive,chen-etal-2023-travel}. Detailed definitions are in Appendix~F.

\subsection{Overall Performance Evaluation}

We conduct extensive experiments to assess the effectiveness of the proposed ALEX framework across five datasets--MQuAKE-CF-3K, MQuAKE-CF-3K-v2, MQuAKE-T, MQuAKE-2002, and MQuAKE-Hard and three representative backbone families: DeepSeek, LLaMA 3.1 8B, and LLaMA 2 7B. As summarized in Table~\ref{mainResults}, ALEX consistently improves both MultiHop-ACC and HopWise-ACC across all model-dataset combinations.

\paragraph{DeepSeek setting.} Under the DeepSeek backbone, ALEX yields substantial gains. On MQuAKE-CF-3K, MeLLo(ALEX) outperforms the vanilla MeLLo by +4.7\% MultiHop-ACC and +7.3\% HopWise-ACC, while PokeMQA(ALEX) achieves even larger improvements of +9.43\% and +13.77\%, respectively. Notably, on the more challenging MQuAKE-Hard dataset, PokeMQA(ALEX) achieves a +35.17\% increase in HopWise-ACC, highlighting ALEX’s advantage in long and difficult multi-hop scenarios.

\paragraph{LLaMA 3.1 8B setting.} ALEX remains effective when applied to larger models. On MQuAKE-T, MeLLo(ALEX) achieves +7.02\% and +19.72\% improvements in MultiHop-ACC and HopWise-ACC, respectively. On MQuAKE-CF-3K-v2, PokeMQA(ALEX) surpasses its baseline by +5.47\% in MultiHop-ACC and +5.78\% in HopWise-ACC, showing the framework’s transferability across scales and benchmarks.

\paragraph{LLaMA 2 7B setting.} Even in more resource-constrained environments, ALEX consistently brings improvements. For instance, on MQuAKE-T, MeLLo(ALEX) improves MultiHop-ACC from 59.26\% to 68.2\% (+15.1\% relative gain) and HopWise-ACC from 32.95\% to 42.63\%. These results confirm ALEX’s scalability and robustness across varying model capacities.

\paragraph{Nuanced Performance on Smaller Models.}
Interestingly, a closer analysis of smaller models like LLaMA 2 7B reveals a nuanced trade-off. While ALEX substantially improves the reliability of the reasoning chain (HopWise-ACC), in some cases, the final answer accuracy (MultiHop-ACC) sees a marginal change. For example, on the MQuAKE-CF-3K dataset, applying ALEX to PokeMQA with LLaMA 2 7B improved HopWise-ACC from 22.91\% to 36.80\%, while MultiHop-ACC experienced a slight decrease (34.20\% → 33.43\%).

This pattern suggests that ALEX’s hierarchical retrieval is highly effective at identifying the correct evidence for each reasoning step. However, smaller models may occasionally struggle to synthesize this correctly retrieved information into a perfect final answer. This indicates that while ALEX successfully fortifies the reasoning process, the final synthesis step remains a bottleneck for less powerful backbone models, a challenge that warrants further investigation.

\paragraph{Performance Across Reasoning Depths}

\begin{figure}[ht]
\centering
\includegraphics[width=0.9\columnwidth]{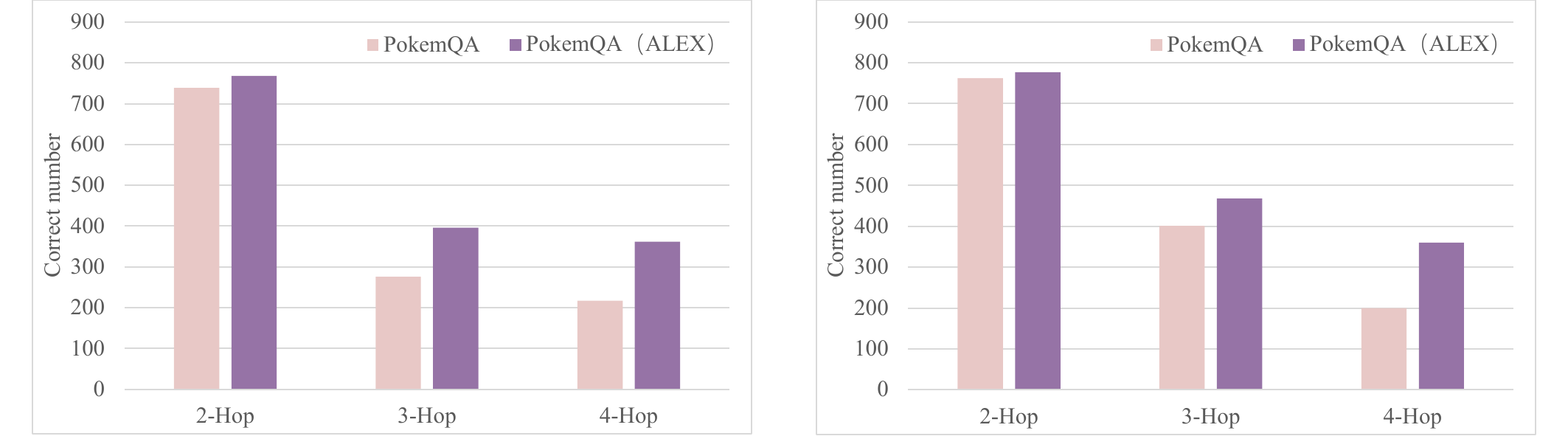} 
  \caption{The comparison of the correct number of PokeMQA and PokeMQA (ALEX) under 2-Hop, 3-Hop, and 4-Hop conditions on the MQuAKE-CF-3K (left) and MQuAKE-CF-3K-v2 (right) datasets.}
\label{difHop_3kandv2}
\end{figure}
To analyze the impact of ALEX under varying reasoning complexities, we compare PokeMQA and PokeMQA(ALEX) on MQuAKE-CF-3K and MQuAKE-CF-3K-v2 under 2-hop, 3-hop, and 4-hop conditions, as illustrated in Figure~\ref{difHop_3kandv2}. ALEX yields progressively greater improvements as reasoning depth increases. In particular, the gains under 4-hop settings are the most pronounced, indicating that ALEX significantly enhances the model’s ability to maintain long-range logical coherence--an area where conventional QA models often struggle.

\paragraph{Component-wise Validation: Cluster and Retrieval Accuracy}

\begin{figure}[ht]
\centering
\includegraphics[width=0.9\columnwidth]{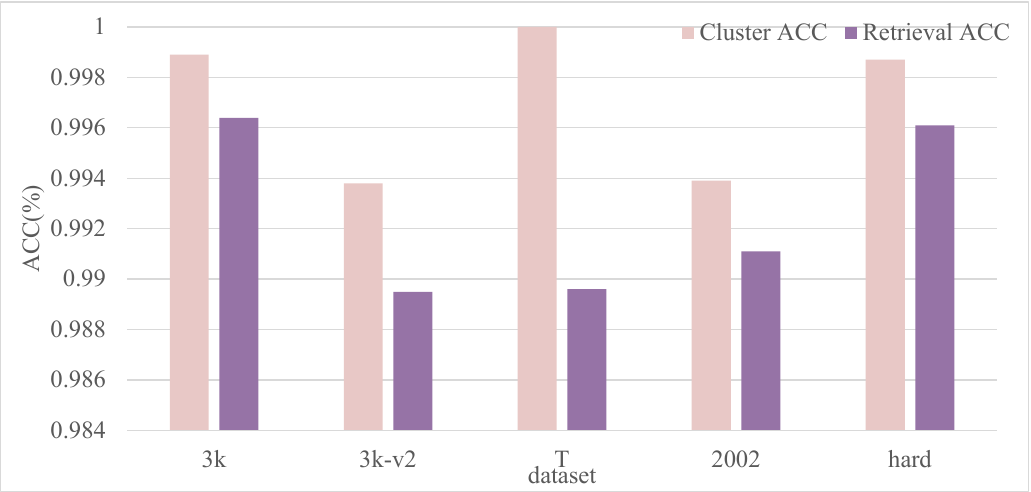} 
  \caption{Performance of the clustering and retrieval components across five datasets. The number of clusters (K=12) was optimized for these datasets as detailed in Appendix B.}
\label{CRACC}
\end{figure}

To validate our two-stage retrieval pipeline, we evaluated its coarse-grained clustering and fine-grained retrieval components. As shown in Figure~\ref{CRACC}, the high accuracy for both stages across all datasets confirms the effectiveness of our design for multi-hop inference.

\paragraph{Summary and Insights}
 
Overall, ALEX demonstrates substantial improvements across the vast majority of our experimental settings, with particularly strong gains on harder questions and longer reasoning chains. The improvements in HopWise-ACC are especially noteworthy, highlighting that ALEX not only helps models reach the correct final answer but also enhances the reliability of intermediate reasoning steps. While performance gains are universal with the powerful DeepSeek backbone, we observed a nuanced trade-off in smaller models like Llama 2 7B, where significant improvements in reasoning-path accuracy (HA) sometimes coincided with a marginal decrease in final-answer accuracy (MA), pointing to a complex interplay between retrieval architecture and model capacity.

\subsection{Ablation Study}
\label{sec:ablation} 
\setlength{\tabcolsep}{4pt} 

\begin{table}[t]
\centering
\small
\begin{tabular}{ccrrrrrr}
\hline
     &                       & \multicolumn{2}{c}{\textbf{M-CF‑3K‑v2}} &
                             \multicolumn{2}{c}{\textbf{M-T}}       &
                             \multicolumn{2}{c}{\textbf{M-Hard}}    \\
\textbf{DEA} & \textbf{IQS} & MA & HA & MA & HA & MA & HA \\
\hline
$\times$ & $\times$ & 36.87 & 30.94 & 70.53 & 59.79 & 62.90 & 57.24 \\
$\surd$  & $\times$ & 48.17 & 42.68 & 82.07 & 71.74 & 67.55 & 62.17 \\
$\times$ & $\surd$  & 41.75 & 35.15 & 75.92 & 64.04 & 74.84 & 69.77 \\
$\surd$  & $\surd$  & \textbf{53.50} & \textbf{47.43} &
                       \textbf{87.33} & \textbf{76.49} &
                       \textbf{79.20} & \textbf{74.35} \\
\hline
\end{tabular}
\caption{Ablation study of the DEA and IQS modules on three MQuAKE datasets (M-CF-3K-v2, M-T, and M-Hard), showing their individual and combined contributions to performance.}
\label{mainAblation}
\end{table}

\begin{figure}[ht]
\centering
\includegraphics[width=0.9\columnwidth]{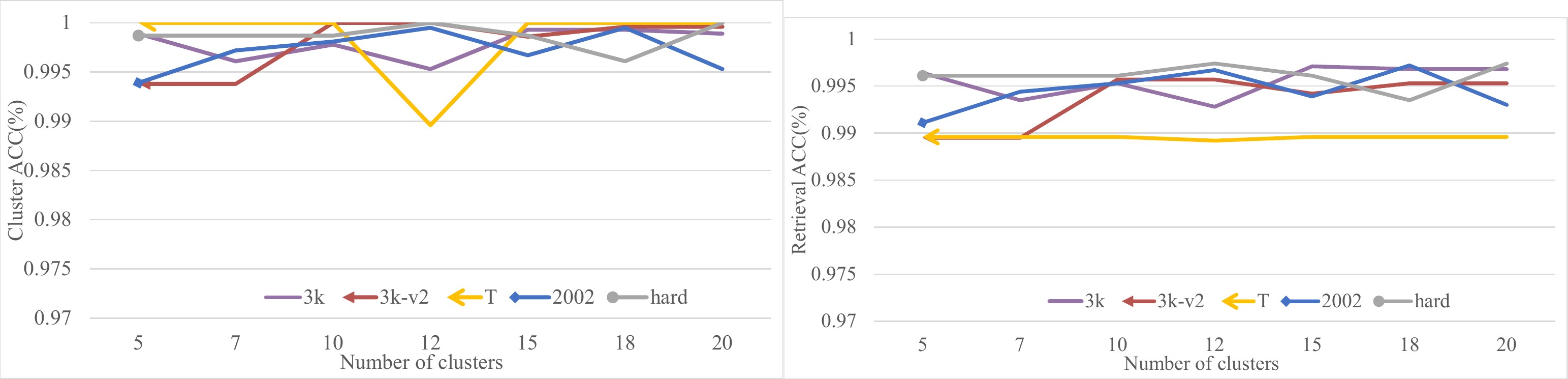}
  \caption{The figure illustrates Cluster ACC and Retrieval ACC under varying numbers of cluster classifications.}
\label{clusterAblation}
\end{figure}

To understand the contribution of individual components in ALEX, we conduct two sets of ablation studies. The first focuses on the effects of our key modules, IQS and DEA, while the second analyzes the robustness of ALEX’s clustering-based retriever.

\paragraph{Impact of IQS and DEA Modules}
We evaluate the role of the IQS and DEA modules by selectively disabling them on three representative datasets. As summarized in Table~\ref{mainAblation}, removing either component causes a substantial performance drop. For instance, on MQuAKE-T, disabling both reduces MultiHop-ACC from 87.33\% to 70.53\%, confirming they make distinct and complementary contributions. Interestingly, the DEA mechanism shows stronger influence on MQuAKE-CF-3K-v2, while IQS exerts more impact on MQuAKE-Hard. This suggests that DEA excels at resolving ambiguities in shallow queries, while IQS better supports deeper entity disambiguation.

\paragraph{Retrieval Effectiveness Across Clustering Granularity}
We further investigate how the retriever’s clustering granularity affects its performance. To analyze robustness, we manually varied the number of clusters (K), deviating from the automatically selected value, and measured performance. This analysis demonstrates ALEX's stability across a range of non-optimal configurations. In Figure~\ref{clusterAblation}, we report Cluster ACC and Retrieval ACC under different K values. Results show that both metrics remain stable across different granularities, with optimal performance typically achieved when K is around 12. This validates the reliability of ALEX's coarse-to-fine retrieval strategy.

\subsection{Efficiency Analysis}

\begin{figure}[t]
\centering
\includegraphics[width=0.9\columnwidth]{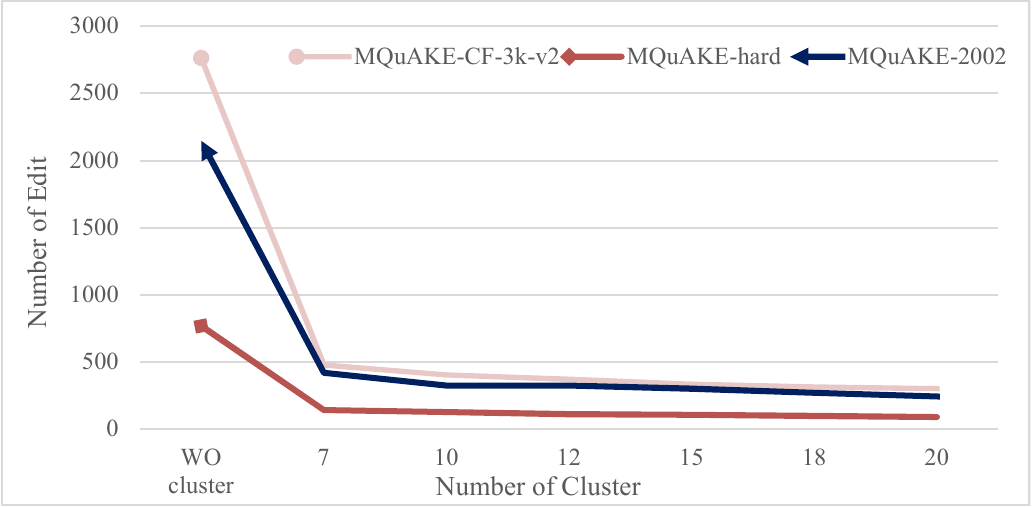}
  \caption{The figure depicts the number of edits that need to be retrieved per question on average for each dataset under different numbers of cluster classifications.}
\label{diffClusterEdit}
\end{figure}

Building on the previous analysis of clustering granularity, we now assess retrieval efficiency under the same varying cluster configurations (K = 7, 10, 12, 15, 18, and 20). This analysis quantifies the reduction in the edit search space, with results presented in Figure~\ref{diffClusterEdit}.

As shown, pre-clustering significantly reduces the number of retrieval candidates compared to a baseline without clustering, which must search the entire edit set. For instance, on MQuAKE-CF-3K-v2, the average retrieval count drops from 2764 to 368 when using 12 clusters—an 86.7\% reduction in the search space. We observe a diminishing return as K increases: moving from 7 to 12 clusters yields large efficiency gains, while further increasing K leads to smaller improvements. This is expected, as finer clustering risks over-fragmentation.

This analysis, combined with the robustness results from our ablation study (Section~\ref{sec:ablation}), confirms that ALEX's coarse-to-fine retrieval is highly effective. It simultaneously achieves significant computational savings—reducing the search space by up to 86.7\%—while maintaining stable and high retrieval accuracy across various cluster granularities. This synergy between efficiency and robustness highlights the practical value of our approach.
\section{Conclusion}
We introduce ALEX, a lightweight framework that addresses the scalability and accuracy limitations of knowledge editing in multi-hop reasoning. ALEX's core innovation is a hierarchical memory structure that reorganizes edits into semantic clusters, reducing retrieval complexity from O(N) to a highly scalable O(K+N/C). Complemented by inferential query synthesis and dynamic evidence adjudication modules, ALEX achieves state-of-the-art performance on MQuAKE datasets, significantly improving reasoning accuracy and reducing the search space by over 80\%. While acknowledging a performance trade-off on smaller models, future work will focus on enhancing its query synthesis module. ALEX presents a promising path toward scalable, efficient, and accurate knowledge editing for large language models.

\section{Limitations And Future work}
We acknowledge certain limitations. A nuanced performance trade-off was observed on smaller models, where improvements in reasoning-path fidelity (HopWise-ACC) did not always translate to equivalent gains in final-answer accuracy (MultiHop-ACC) . This suggests a potential bottleneck in the synthesis capacity of these models when processing the retrieved facts. Furthermore, the Inferential Query Synthesis (IQS) module, while effective, can be susceptible to ambiguous inputs.

Future work will focus on enhancing the IQS module through techniques like edit-conditioned prefix tuning. Regarding efficiency, this study prioritized validating the algorithmic approach. For the sake of experimental rigor, our subsequent work will involve developing a C++ implementation of ALEX. This will facilitate a direct and equitable performance benchmark against highly-optimized, low-level libraries like Faiss.

\section*{Acknowledgements}
This research is partially supported by The National Social Science Fund of China No. 18ZDA200, Introducing Talents of Studying Overseas Fund of Hebei No. C20230339, Special Science and Technology Fund of Hebei Normal University No. L2023T03.

\bibliography{custom}

\appendix
\section{DEA Engine Formulation}
\label{app:dea}
This appendix details the mathematical formulation of the Dynamic Evidence Adjudication (DEA) Engine, as referenced in Section 3.4.

\subsection{Cluster-Level Statistical Filtering}
Given a query embedding $\phi(q)$ and a set of K cluster centroids $\{\mu_{i}\}_{i=1}^{K}$, we first compute the cosine similarity score $s_i = \cos(\phi(q), \mu_i)$ for each cluster. These scores are then normalized using a z-score transformation to identify statistically relevant clusters:

\begin{equation}
  z_{i} = \frac{s_{i} - \bar{s}}{\sigma_{s}}.
\end{equation}

Here, $\bar{s}$ and $\sigma_{s}$ are the mean and standard deviation of the similarity scores over all clusters. Clusters with $z_{i} \ge \zeta$ (where $\zeta=1.0$) are retained as candidates, up to a maximum of $M=3$.

\subsection{Edit-Level Semantic Adjudication}
Within the filtered clusters, each candidate edit $e_j$ is scored using a composite function. This function integrates direct evidence (query-edit similarity) and inferential evidence (query similarity to hypothetical questions $\mathcal{H}(e_j)$). Consistent with the implementation described in Section 3.4, we use max-pooling for aggregation:

\begin{equation}
  \Psi(e_{j}) = \alpha \cdot \cos(\phi(q), \phi(e_{j})) + \beta \cdot \max_{h \in \mathcal{H}(e_{j})} \cos(\phi(q), \phi(h)).
\end{equation}

The edit with the highest adjudication score $e^{*} = \arg\max_{e_{j}} \Psi(e_{j})$ is selected. The hyperparameters $\alpha$ and $\beta$ are both set to 0.5.

\section{Clustering Mechanism Details}
\label{app:clustering}
This section provides technical details for the Semantic Manifold Partitioning (SMP) engine's clustering mechanisms.

\subsection{Automatic Cluster-Count Selection}
To determine the optimal number of clusters $K$ automatically, we use a method that balances cluster cohesion (measured by silhouette score $S(K)$) and separation (measured by the elbow gap $E(K)$ based on within-cluster variance $W(K)$). The chosen cluster count $K^*$ is the one that maximizes the following objective:
\begin{equation}
  K^* = \arg\max_{K} \left[ \alpha\,S(K) - \beta\,E(K) \right].
\end{equation}

For the experiments in this paper, this procedure resulted in an optimal count of \textbf{K=12}.

\subsection{Dynamic Manifold Adaptation}
To ensure the manifold structure adapts to evolving data, the SMP engine incorporates a dynamic adaptation mechanism as described in Section 3.2. This mechanism monitors cluster quality and triggers a partial reclustering of the most affected clusters if either of the following conditions is met:
\begin{itemize}
    \item The average silhouette score of any single cluster falls below a threshold $\theta_{s} = 0.5$.
    \item The global average silhouette score drops by more than 20\% relative to its post-training peak.
\end{itemize}

\section{IQS Module Implementation Details}
\label{app:iqs}
This appendix details the implementation of the Inferential Query Synthesis (IQS) module.

\subsection{Hypothetical Question Generation}
For each factual edit $e_j$, we use \textbf{GPT-3.5-turbo} (with temperature = 0.7) to generate $N_h=3$ diverse, plausible hypothetical questions. The prompts are designed to rephrase the declarative fact into natural-sounding questions. Example prompts and resulting question-fact pairs are shown in Table~\ref{tab:IQS-prompt}.

\begin{table}[h!]
\centering
\small
\begin{tabular}{|p{0.9\linewidth}|}
\hline
\textbf{Instruction Prompt Template} \\
\hline
Given the following fact, please generate three different questions that this fact could answer. The questions should be natural and what a curious person might ask.
\newline\newline
\textbf{Fact}: "The Eiffel Tower is located in Paris."
\newline\newline
\textbf{Generated Hypothetical Questions}:
\begin{enumerate}
    \item Where is the Eiffel Tower located?
    \item In which city can you find the Eiffel Tower?
    \item What is a major landmark in Paris?
\end{enumerate} \\
\hline
\end{tabular}
\caption{An example of the prompt template used for the IQS module and the resulting hypothetical questions for a given fact.}
\label{tab:IQS-prompt}
\end{table}

\subsection{Caching and Quality Control}
A persistent caching layer is used to store generated questions, minimizing redundant API calls. To ensure semantic fidelity, we apply filtering heuristics: we discard questions with fewer than 3 tokens or no named entities, and require at least 60\% n-gram overlap with the original edit.

\section{Details of MQuAKE-related Datasets}
\label{app:datasets}
Experiments were conducted on five MQuAKE-related datasets, which are designed for evaluating knowledge editing in the context of multi-hop question answering. Table~\ref{tab:dataset_stats} summarizes the distribution of questions by reasoning depth (hop count) across these datasets.
\begin{table}[h!]
\centering
\small
\begin{tabular}{lrrrr}
\hline
& \multicolumn{4}{c}{\textbf{Hop Count}} \\
\textbf{Dataset} & \textbf{2-Hop} & \textbf{3-Hop} & \textbf{4-Hop} & \textbf{Total} \\
\hline
MQuAKE-CF-3K & 1135 & 1136 & 729 & 3000 \\
MQuAKE-CF-3K-v2 & 1135 & 1136 & 729 & 3000 \\
MQuAKE-T & 1421 & 445 & 2 & 1868 \\
MQUAKE-2002 & 966 & 625 & 411 & 2002 \\
MQUAKE-Hard & 0 & 0 & 429 & 429 \\
\hline
\end{tabular}
\caption{The number of questions with different hop counts and the total number in each MQuAKE-related dataset.}
\label{tab:dataset_stats}
\end{table}

\begin{itemize}
    \item \textbf{MQUAKE-CF-3K} and its updated version \textbf{-V2} contain 3,000 multi-hop questions with counterfactual edits derived from Wikidata.
    \item \textbf{MQUAKE-T} contains 1,868 instances focused on temporal knowledge updates, using facts from different Wikidata dumps.
    \item \textbf{MQUAKE-2002} is a filtered, knowledge-conflict-free set of 2,002 questions for evaluating pure editing performance.
    \item \textbf{MQUAKE-Hard} is a challenging subset of 429 questions, each involving four edited facts, testing the limits of multi-hop reasoning with edited knowledge.
\end{itemize}

\section{Experimental Setup}
\label{app:exp_setup}
All experiments were conducted on a machine with an Intel Core i9-13900K processor, 128GB of RAM, and two NVIDIA RTX 4090 GPUs. We used Python 3.9 and PyTorch 2.0.

Our training data is sourced from the prior work of PokeMQA, which was generated by Vicuna 13B. The dual-objective MPNet encoder was initialized from Sentence Transformers library's pretrained weights and optimized using AdamW (learning rate $2e-5$, weight decay $0.01$) with a linear warmup for the first $10\%$ of steps. We employed an early stopping mechanism based on the Yield Detection Index (YDI) on a validation set to prevent overfitting. Baselines like MeLLo and PokeMQA were reimplemented using their official configurations for a fair comparison.

\section{Evaluation Metrics}
\label{app:metrics}
We use four primary metrics to evaluate performance from different perspectives.

\paragraph{Multi-hop Accuracy (MultiHop-ACC)} Measures the percentage of questions for which the final generated answer exactly matches the gold answer.
\begin{equation}
 \text{MultiHop-ACC} = \frac{1}{N}\sum_{i=1}^{N} \mathbf{1}(o_{n}^{(i)} = o_{n}^{*(i)}).
\end{equation}

\paragraph{Hop-wise Answering Accuracy (Hop-Acc)} A stricter metric that measures the percentage of questions where the model's entire reasoning path (all intermediate entities and answers) exactly matches the gold path.
\begin{equation}
 \text{Hop-Acc} = \frac{1}{N}\sum_{i=1}^{N} \mathbf{1}(\mathcal{P}^{(i)} = \mathcal{P}^{*(i)}).
\end{equation}

\paragraph{Cluster Accuracy (Cluster-Acc)} Evaluates the retriever's first stage, measuring the percentage of times the correct cluster (containing the gold edit) is selected.
\begin{equation}
 \text{Cluster-Acc} = \frac{1}{N}\sum_{i=1}^{N} \mathbf{1}(e^{*(i)} \in \mathcal{C}^{(i)}).
\end{equation}

\paragraph{Retrieval Accuracy (Retrieval-Acc)} Measures the end-to-end success of the retriever, counting the percentage of times the final selected edit is the correct gold edit.
\begin{equation}
 \text{Retrieval-Acc} = \frac{1}{N}\sum_{i=1}^{N} \mathbf{1}(e^{*(i)} \in \mathcal{E}^{(i)} \land \mathcal{E}^{(i)} \subseteq \mathcal{C}^{(i)}).
\end{equation}

\section{Complexity Justification for Hierarchical Retrieval}
\label{app:complexity}
This section provides a formal justification for the retrieval complexity reduction from $O(N)$ to $O(K + N/C)$.

Let $N$ be the total number of edits, $K$ be the number of clusters, and $C \approx N/K$ be the average number of edits per cluster. The retrieval process consists of two steps:
\begin{enumerate}
    \item \textbf{Cluster-level filtering}: The query is compared against all $K$ cluster centroids. This has a complexity of $O(K)$.
    \item \textbf{Edit-level scoring}: A small, constant number of clusters $M$ (empirically $M \le 3$) are selected. Scoring is performed only on the edits within these clusters, leading to a complexity of $O(M \cdot C) \approx O(N/K)$.
\end{enumerate}

The total complexity is the sum of these two steps:
\begin{equation}
  \text{Total Complexity} = O(K) + O(N/K) = O(K + N/C).
\end{equation}
This hierarchical approach is significantly more scalable than a flat search of $O(N)$, which is both theoretically sound and empirically validated in our experiments.

\end{document}